\documentclass[conference]{IEEEtran}
\IEEEoverridecommandlockouts
\usepackage{cite}
\usepackage{amsmath,amssymb,amsfonts}
\usepackage{algorithmic}
\usepackage{graphicx}
\usepackage{textcomp}
\usepackage{xcolor}

\usepackage{multirow}
\usepackage{multicol}
\usepackage{booktabs}
\usepackage{hyperref}
\usepackage[subrefformat=parens]{subcaption}

\newcommand{\bn}[1]{\textbf{#1}}
\newcommand{\NA}{\multicolumn{1}{c}{---}}
\newcommand{\NAR}{\multicolumn{1}{c|}{---}}

\def\BibTeX{{\rm B\kern-.05em{\sc i\kern-.025em b}\kern-.08em
    T\kern-.1667em\lower.7ex\hbox{E}\kern-.125emX}}
\begin{document}

\title{Program Repair with Minimal Edits Using CodeT5
\thanks{This work was supported by the Japan Society for the Promotion of Science (JSPS) KAKENHI Grant Number JP23H03508.}
}

\author{

\IEEEauthorblockN{Atsushi Shirafuji}
\IEEEauthorblockA{
\textit{University of Aizu}\\
Aizu-Wakamatsu, Japan \\
m5261161@u-aizu.ac.jp
}

\and

\IEEEauthorblockN{Md. Mostafizer Rahman}
\IEEEauthorblockA{
\textit{Dhaka University of}\\
\textit{Engineering \& Technology}\\
Gazipur, Bangladesh\\
mostafiz26@gmail.com
}

\and

\IEEEauthorblockN{Md Faizul Ibne Amin}
\IEEEauthorblockA{
\textit{University of Aizu}\\
Aizu-Wakamatsu, Japan \\
aminfaizul007@gmail.com
}

\and

\IEEEauthorblockN{Yutaka Watanobe}
\IEEEauthorblockA{
\textit{University of Aizu}\\
Aizu-Wakamatsu, Japan \\
yutaka@u-aizu.ac.jp
}
}

\maketitle

\begin{abstract}
Programmers often struggle to identify and fix bugs in their programs.
In recent years, many language models (LMs) have been proposed to fix erroneous programs and support error recovery.
However, the LMs tend to generate solutions that differ from the original input programs.
This leads to potential comprehension difficulties for users.

In this paper, we propose an approach to suggest a correct program with minimal repair edits using CodeT5.
We fine-tune a pre-trained CodeT5 on code pairs of wrong and correct programs and evaluate its performance with several baseline models.

The experimental results show that the fine-tuned CodeT5 achieves a pass@100 of 91.95\% and an average edit distance of the most similar correct program of 6.84, which indicates that at least one correct program can be suggested by generating 100 candidate programs.
We demonstrate the effectiveness of LMs in suggesting program repair with minimal edits for solving introductory programming problems.
\end{abstract}


\section{Introduction}
\label{sec:intro}

Most of the time, programmers edit source code to fix bugs, add new features, or change existing features.
Recent studies have shown that these edits are repetitive~\cite{nguyen2013study,chakraborty2022codit}, and manually repeating the edits can be error-prone and time-consuming~\cite{ray2013detecting}.
Many language models (LMs) have been proposed to assist debugging by identifying errors and suggesting fixes in source code.
Suggesting correct code for wrong code using LMs can significantly reduce these repetitive edits and reduce the effort programmers spend to correct the wrong code. 

Several deep learning models, such as recurrent neural network (RNN), long short-term memory (LSTM), bidirectional LSTM (BiLSTM), LSTM with an attention mechanism (LSTMAttn), and Transformer, are used to correct erroneous source code and provide various learning supports, such as predicting the next tokens~\cite{svyatkovskiy2020intellicode, terada2021completion}, fixing bugs in wrong code~\cite{berabi2021tfix, rahman2021repair, matsumoto2021repair, joshi2022repair}, and improving the performance and readability~\cite{madaan2023improving, madaan2023self-refine}.
In particular, LMs based on the Transformer architecture~\cite{vaswani2017transformer} have demonstrated exceptional performance across various tasks, not only in texts~\cite{devlin2019bert, brown2020gpt3} but also in code~\cite{chen2021codex, li2022alphacode, wang2021codet5, fried2023incoder, nijkamp2022codegen, zheng2023codegeex, li2023starcoder, luo2023wizardcoder}, images~\cite{dosovitskiy2021vit}, and videos~\cite{arnab2021vivit}.

\begin{figure}
    \captionsetup{belowskip=-12pt}
    \centerline{\includegraphics[width=.9\linewidth]{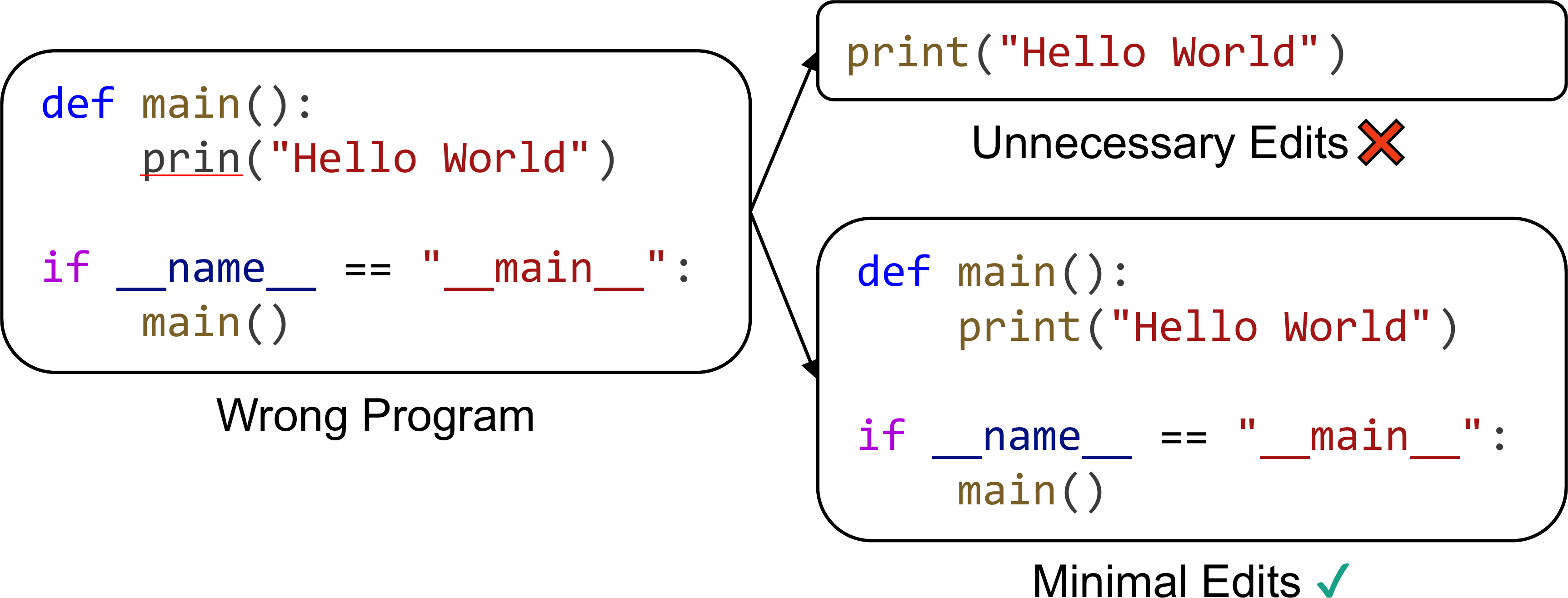}}
    \caption{\textbf{Motivating example of program repair.}}
    \label{fig:problem}
\end{figure}
However, the code generated by LMs often deviates from user expectations, necessitating additional effort to edit to rectify the code.
As illustrated in Figure~\ref{fig:problem}, although the top-right program is a more popular solution than the bottom-right program, it requires larger edits from the wrong program.
Since it can confuse the user in understanding the suggested repair, suggesting a program with minimal edits is desirable.
Despite the importance of whether the generated programs are helpful for learners, to the best of our knowledge, no previous work has been undertaken to investigate syntactic similarity and functional correctness simultaneously. 

In this work, to mitigate the learners' burden of finding the bugs and support more effective learning programming, we propose using an LM trained on source code, CodeT5~\cite{wang2021codet5}, to suggest a correct program with minimal edits to repair the given wrong program.
By suggesting the correct program that is more aligned with the user-written program, we can avoid confusing users due to the large difference in the suggested program from the original program.

We fine-tune a pre-trained CodeT5 on a dataset of code pairs consisting of wrong and correct programs collected from Aizu Online Judge (AOJ)~\cite{watanobe2004aoj}.
We evaluate the performance of the fine-tuned CodeT5 compared with the following baseline models.
\begin{itemize}
    \item Naive copy model to copy the input program as the output program.
    \item Naive retrieval model to retrieve the most similar program from the training data.
    \item Sequence-to-sequence (Seq2Seq) model consisting of a BiLSTM and an LSTMAttn.
\end{itemize}

Our experimental results demonstrate that the fine-tuned CodeT5 achieves a 91.95\% on pass@100.
It indicates that at least one correct program can be suggested by generating 100 candidate programs for 91.95\% of the wrong programs.
Moreover, the edit distance of the most similar correctly generated program is 6.84 on average, whereas the human-crafted repair has 10.76 on average.
We show that the fine-tuned CodeT5 can generate correct programs for wrong programs with minimal edits to assist in solving introductory programming problems.

The contributions of this work are as follows.
\begin{itemize}
    \item We demonstrate that fine-tuning CodeT5 on code pairs of (wrong, correct) programs performs well on program repair. 
    \item We show that the fine-tuned CodeT5 can generate repaired programs with minimal edits compared to human-crafted repair.
\end{itemize}

\section{Related Work}
\label{sec:related-work}

Automated program repair (APR) has been a subject of increasing interest with the growth of software systems, aiming to reduce the time and effort spent on debugging by programmers.
Many studies have utilized deep learning techniques, such as RNNs and Transformers, considering the task of converting the wrong program into a correct one, similar to the neural machine translation task.

As the earlier works leveraged RNNs for program repair, using Seq2Seq based on RNN~\cite{bhatia2016repair, pu2016skp, tufano2018repair}, LSTM~\cite{chen2021sequencer, matsumoto2021repair}, BiLSTM~\cite{rahman2021repair, rahman2023codeBiLSTM}, and LSTMAttn~\cite{hata2019repair}, has been proposed.
For other attempts, \textsc{Hoppity}~\cite{dinella2020hoppity} used the graph neural network to capture the graph structure, and \textsc{CoCoNuT}~\cite{lutellier2020coconut} used the convolutional neural network instead of an LSTM to model source code at different granularity levels.
As one of the works considering the edit distance of generated programs, Gulwani~et~al.~\cite{gulwani2018repair} proposed \textsc{Clara} using the syntactic difference (tree-edit-distance) as the cost function to find the program repair from the existing correct student solutions for introductory programming education.
Similarly, Lu~et~al.~\cite{lu2021fapr} proposed a fast and accurate program repair tool, \textsc{FAPR}, that outperformed \textsc{Clara} in suggesting the correct and smaller programs according to the qualitative evaluation.
As an application, Parihar~et~al.~\cite{parihar2017repair} applied program repair, enabling automatic grading of incorrect submissions that contain syntax errors using test cases and awarding partial marks for them.

In recent years, Transformer-based approaches have performed remarkably well and are now a dominant model.
Especially, Transformer-based~\cite{drain2021deepdebug, lu2021codexglue}, BERT~\cite{devlin2019bert}-based~\cite{feng2020codebert, lu2021codexglue}, and T5~\cite{raffel2020t5}-based~\cite{berabi2021tfix, wang2021codet5} models are proposed.
CodeT5~\cite{wang2021codet5} has demonstrated the capability in program repair on the CodeXGLUE benchmark~\cite{lu2021codexglue}.

More recently, using large language models (LLMs) trained on source code has shown the capability in APR.
Several works~\cite{prenner2021repair, joshi2022repair, zhang2022repair, jin2023inferfix} proposed APR systems leveraging LLMs such as Codex~\cite{chen2021codex}.
Not only fixing syntactic or semantic bugs, but Codex has also shown the ability to fix security bugs~\cite{pearce2021securitybugs}, improving time performance~\cite{madaan2023improving} and code readability~\cite{madaan2023self-refine}.

Our approach shares similarities with the works mentioned above.
However, it differs in focusing on minimal edits for program repair to better align with the user-written programs.

\section{Experiments}
\label{sec:experiments}

\subsection{Proposed Approach}
\label{sec:experiments:approach}

\begin{figure}
    \captionsetup{belowskip=-12pt}
    \centerline{\includegraphics[width=\linewidth]{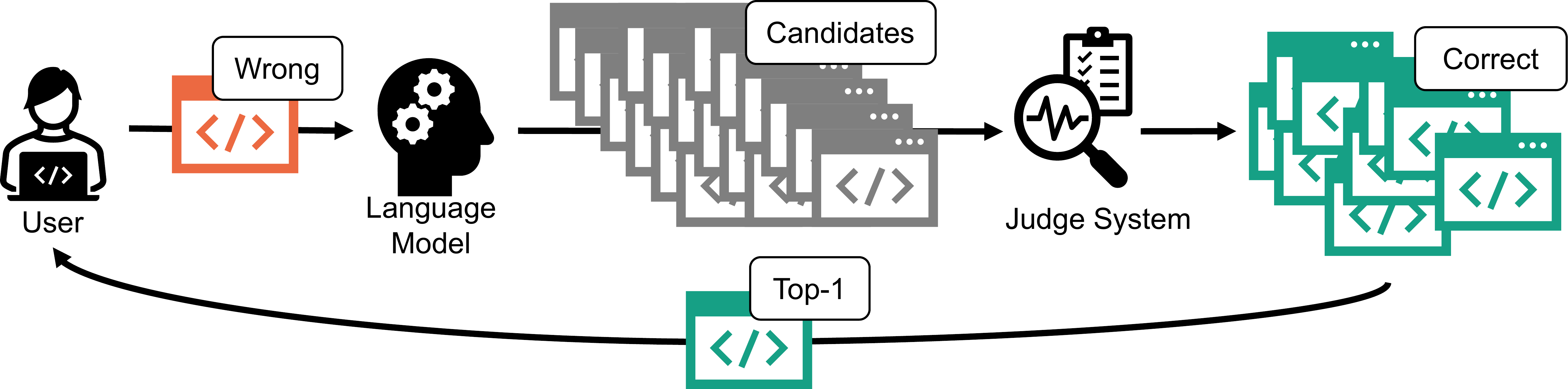}}
    \caption{\textbf{Illustration of the proposed approach.}}
    \label{fig:approach}
\end{figure}

The proposed approach is illustrated in Figure~\ref{fig:approach}.
A user inputs a wrong program, failing to solve a programming problem.
After an LM generates multiple candidate programs, a judge system validates the functional correctness of the generated programs.
For the correctly generated programs, the most similar program, i.e., the program with the smallest edit distance with the input program, is suggested to the user.
This process allows the user to obtain \textit{a repaired program with minimal edits.}

Note that, in this work, we also use naive models instead of the LM for comparison.
In addition, in the experiments, we generate 100 candidate programs.
Increasing the number of candidate programs enhances the accuracy of suggesting correct programs as it can search for better programs.
However, the system response time must be increased as the number of candidate programs increases since LMs require much computational time for inference.

\subsection{Dataset}
\label{sec:experiments:dataset}

We use Python 3 programs submitted on AOJ~\cite{watanobe2004aoj, watanobe2022aoj} for the dataset to train and evaluate the models.
We target a set of programming problems named \textit{Introduction to Programming I} (ITP1)\footnote{\url{https://onlinejudge.u-aizu.ac.jp/courses/lesson/2/ITP1/all}.}, an introductory course with 44 programming problems.
The course is designed for introduction to programming and ranges from requiring standard input and output to class and method definitions.

\begin{figure}[h]
    \centerline{\includegraphics[width=.55\linewidth]{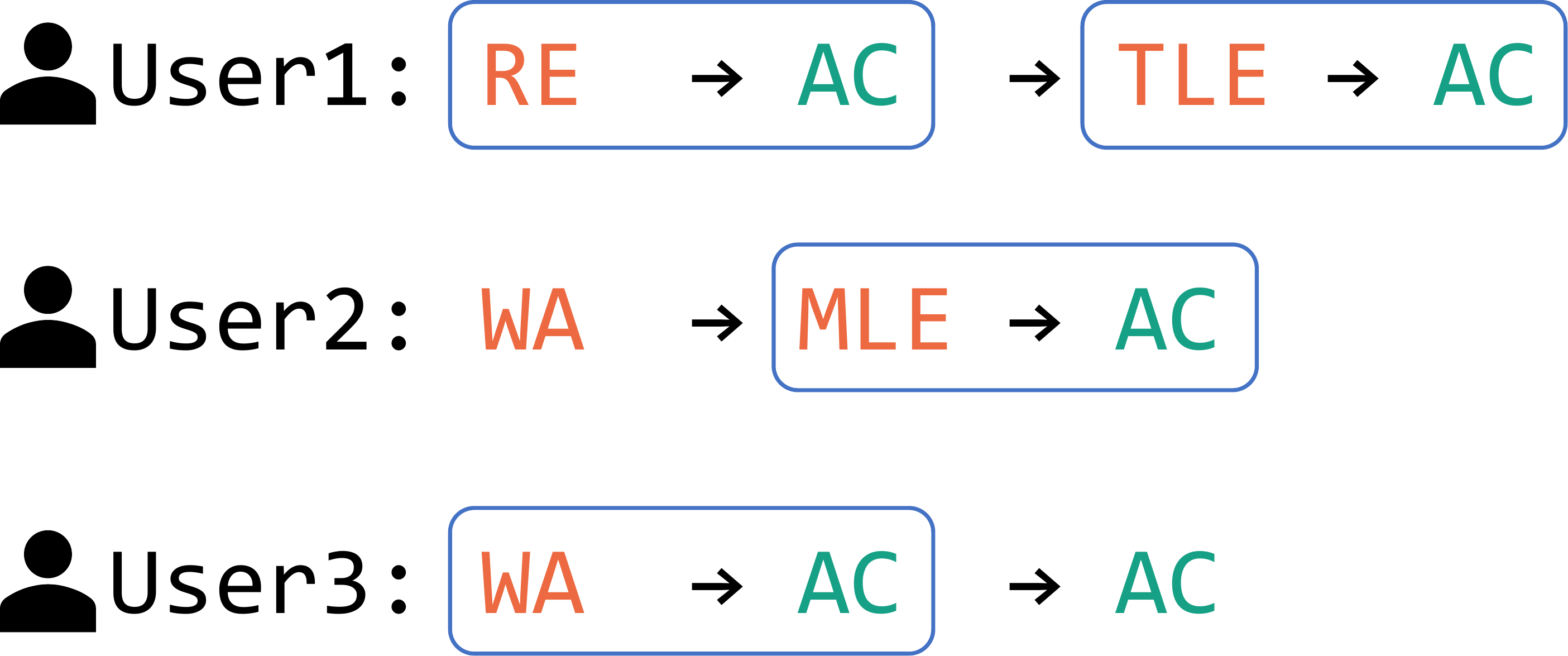}}
    \caption{\textbf{Illustration of collecting code pairs consisting of wrong and correct programs from the same user.} \textit{AC} indicates correct programs, and \textit{WA}, \textit{RE}, \textit{TLE}, and \textit{MLE} indicate wrong programs, such as wrong answer, runtime error, time limit exceeded, and memory limit exceeded, respectively.}
    \label{fig:code_pairs}
\end{figure}

For the program repair task, we collect code pairs of wrong and correct programs from AOJ, as shown in Figure~\ref{fig:code_pairs}.
If a wrong program is submitted before the correct program, we consider it \textit{an attempt} and make a code pair (wrong, correct) with the correct program.
For program consistency, each code pair consists of the submissions from the same user.

As preprocessing, we only use the code whose token-based length is more than 0 and less than 256.
In addition, we remove duplicated code pairs to avoid overfitting and cheating in training and evaluating models.

After shuffling the collected code pairs, we split the code pairs into 90\% for training or fine-tuning, 5\% for validation, and 5\% for testing.
Table~\ref{table:code_pairs} shows the number of code pairs and their average edit distance between wrong and correct programs.

\begin{table}
    \centering
    \caption{The number of code pairs in each set.}
    \label{table:code_pairs}
    \begin{tabular}{lcc}
        \toprule
        & \#Code Pairs & Avg. Edit Distance (Std) \\
        \midrule
        Train & 52,526 (90\%) & 10.72 (13.68) \\
        Valid & 2,918 (5\%) & 10.95 (14.39) \\
        Test & 2,918 (5\%) & 11.34 (17.14) \\
        \midrule
        Total & 58,362 & 10.76 (13.91) \\
        \bottomrule
    \end{tabular}\\[-8pt]
\end{table}

\subsection{Evaluation Metrics}
\label{sec:experiments:metrics}

\subsubsection{Pass Rate}
\label{sec:experiments:metrics:pass-rate}

Pass rate (also known as success rate) is a metric showing the percentage of problems solved by generating $k$ programs for each problem~\cite{chen2021codex}.
A problem is considered solved if any generated program solves the problem (i.e., passes all test cases).
However, this work uses this metric to show \textit{the percentage of wrong programs that are repaired by generating $k$ programs for each wrong program.}
Therefore, a wrong program is considered repaired if any generated program solves the problem.

In this work, we use an unbiased estimator of pass rate for programs, pass@$k$, inspired by Chen~et~al.~\cite{chen2021codex}.
Pass@$k$ is denoted as Formula~\ref{formula:pass-at-k}, where $n \geq k$ is the number of samples and $c \leq n$ is the number of correct samples.
\begin{equation}
    \text{Pass@$k$} := \mathop{\mathbb{E}}_{\text{Programs}} \left[ 1 - \frac{{\binom{n-c}{k}}} {\binom{n}{k}} \right]
    \label{formula:pass-at-k}
\end{equation}
Whether the generated program solved the given programming problem (i.e., functional correctness) is validated by executing it using hidden test cases.
More detail about the evaluation of generated programs is described in Section~\ref{sec:experiments:environment}.
We report pass@$k$ at $k \in \{ 1, 10, 100 \}$ where $n = 100$ samples are generated for each wrong program.

\subsubsection{Compilability}
\label{sec:experiments:metrics:compilability}

Compilability is the syntactic correctness of programs, showing whether the program passes the compilation.
It does not account for semantic correctness.
Therefore, a compilable program can be a wrong program.
Inspired by the pass@$k$, we define compilable@$k$ as Formula~\ref{formula:compilable-at-k}, where $n \geq k$ is the number of samples and $c \leq n$ is the number of compilable samples.
\begin{equation}
    \text{Compilable@$k$} := \mathop{\mathbb{E}}_{\text{Programs}} \left[ 1 - \frac{{\binom{n-c}{k}}} {\binom{n}{k}} \right]
    \label{formula:compilable-at-k}
\end{equation}
We report compilable@$k$ at $k \in \{ 1, 10, 100 \}$ where $n = 100$.

\subsubsection{BLEU}
\label{sec:experiments:metrics:bleu}

To evaluate the syntactic similarity of generated programs against the expected correct programs, we report smoothed BLEU-4 scores~\cite{papineni2002bleu, lin2004orange}.
Although several works reported that BLEU is not a good metric in code-related tasks, as it does not account for functional correctness~\cite{ren2020codebleu, roziere2020transcoder, chen2021codex}, we use this metric to show \textit{how much the generated programs syntactically match the expected programs} for reference.
In this work, BLEU scores are computed based on tokens, which are tokenized using the pre-trained byte-pair encoding (BPE) tokenizer of CodeT5, \texttt{codet5-base}\footnote{\url{https://huggingface.co/Salesforce/codet5-base}}, from the tokenizers\footnote{\url{https://github.com/huggingface/tokenizers}.} library.
To compute the BLEU scores, we employ the evaluate\footnote{\url{https://github.com/huggingface/evaluate}.} library.

\subsubsection{Exact Match}
\label{sec:experiments:metrics:exact-match}

Exact Match is a metric that measures the percentage of generated programs that exactly match the expected target programs.
This metric provides insights into the model's ability to replicate the exact solution, which can be particularly useful in certain use cases where the exact replication of the solution is required.
However, it is noteworthy that a low Exact Match score does not necessarily imply poor performance, as the model might generate functionally correct but syntactically different programs.
Therefore, while Exact Match provides valuable information, it should be interpreted with other metrics that measure functional correctness and syntactic similarity, such as pass@$k$, compilable@$k$, and BLEU.
We report the ratio of the Exact Match.

\subsubsection{Edit Distance}
\label{sec:experiments:metrics:edit-distance}

Edit distance (also known as Levenshtein distance~\cite{levenshtein1966}) is used to measure the dissimilarity between the source and the generated programs.
It quantifies the minimum number of character-level changes (insertions, deletions, and substitutions) required to transform the source program into the generated one.
This metric indicates the magnitude of alterations the model makes to obtain the correct program, as well as the number of edits required to repair the wrong program.

In addition to the character-based edit distance, we also report two specific edit distances: correct and top-1.
The correct edit distance is computed between the source and correctly generated programs, indicating the alternations required to correct the wrong program, whereas the default edit distance includes edits from wrong to wrong programs.
The top-1 edit distance, on the other hand, is calculated between the source and the most syntactically similar correct program, providing insight into the best-case scenario of successful program repair.

Unlike other metrics, the edit distance does not involve the target program; the calculation is performed solely using the source and generated programs.
This makes it a helpful metric in understanding the model's strategy in problem-solving - whether it heavily modifies the source program or makes minimal edits.
\textbf{This work primarily focuses on minimizing the edit distance metrics} while keeping the functional correctness of the generated programs.

\subsection{Models}
\label{sec:experiments:models}

\subsubsection{Naive}
\label{sec:experiments:models:naive}

For the simplest baseline models to compare with the fine-tuned LM, we employ two types of naive models: Naive Copy and Naive Retrieval.

\paragraph{Naive Copy}
We refer to the model Naive Copy as the model to copy the input program as the output program.
Since the input program is always wrong, the output program is always wrong (i.e., pass@$k$ is always 0\% at any $k$).
However, note that the output program can be compilable (i.e., compilable@$k$ can be greater than 0\%) since the wrong program includes the wrong answer, which passed the compilation, but the output is wrong.
Although Naive Copy does not repair programs, it can be a base comparison from the perspective of BLEU scores, as it constantly generates a similar program to the target program.

\paragraph{Naive Retrieval}
We refer to the model Naive Retrieval as the model to retrieve the most similar program from the training data in each programming problem using linear search.
The program with the shortest edit distance in the training data is considered the most similar program.
Since Naive Retrieval retrieves a correct program from the training data, programs generated by this model are ensured to be correct (i.e., pass@$k$ and compilable@$k$ are always 100\% at any $k$).
The key aspect of comparing with this model is the edit distance.
This model can result in 100\% correctness but does not necessarily generate the most helpful program.

\subsubsection{Seq2Seq}
\label{sec:experiments:models:seq2seq}

For the baseline model based on LMs, we also use an LSTM-based Seq2Seq model.
A Seq2Seq model is composed of an encoder and a decoder parts.
The encoder utilizes a BiLSTM to parse the input sequences and extract their features.
The decoder, on the other hand, employs an LSTMAttn.
The attention mechanism allows the decoder access to all parts of the input sequences.
The total number of model parameters is 12.4M.
We employ fairseq\footnote{\url{https://github.com/facebookresearch/fairseq}.}~\cite{ott2019fairseq} for model training.

\subsubsection{CodeT5}
\label{sec:experiments:models:codet5}

CodeT5~\cite{wang2021codet5} is a Transformer-based~\cite{vaswani2017transformer} LM specifically tailored for source code.
It is a variant of the T5~\cite{raffel2020t5} model, which is designed to handle any text-to-text conversion tasks.
CodeT5 has shown the capability in the code refinement (program repair) task from the CodeXGLUE benchmark~\cite{lu2021codexglue} by fine-tuning.
We use the \texttt{codet5-base}\footnote{\url{https://huggingface.co/Salesforce/codet5-base}} model, which has 220M parameters, and fine-tune it on our dataset.
We use the transformers\footnote{\url{https://github.com/huggingface/transformers}.} library to load the model and conduct the fine-tuning.

\subsection{Environment}
\label{sec:experiments:environment}

For the evaluation of program correctness, generated programs are executed on an isolated judge system to validate the functional correctness.
The judge system uses hidden test cases for the programming problem provided by AOJ.
The program is judged correct if it passes all hidden test cases and is incorrect otherwise.

Experiments for Seq2Seq and CodeT5 models (especially training and inference) are conducted in a GPU environment with one NVIDIA A100 40GB GPU.
Experiments for naive models and evaluations by the judge systems are conducted in a CPU environment.

\section{Results}
\label{sec:results}

\setlength\tabcolsep{5pt}
\begin{table*}
    \centering
    \caption{\textbf{Results evaluated on the test set.} Naive Copy copies the input data as the output data, which is ensured to be incorrect but can be compilable. Naive Retrieval retrieves the most similar program from the training data, which is ensured to be correct. Seq2Seq and CodeT5 models generate 100 samples for each code pair. \textit{Correct} indicates the average edit distance between input and correctly generated programs, and \textit{Top-1} indicates the average edit distance between input and the most similar correctly generated programs. The best score in LM-based models for each metric is represented in bold.}
    \label{table:results}
    \begin{tabular*}{0.96\linewidth}{c|ccc|ccc|c|c|ccc}
        \toprule
        \multirow{2}{*}{Model} & \multicolumn{3}{c|}{Pass@$k$} & \multicolumn{3}{c|}{Compilable@$k$} & \multirow{2}{*}{BLEU}
            & \multirow{2}{0.04\linewidth}{Exact Match} & \multicolumn{3}{c}{Edit Distance (Std)} \\
        & $k = 1$ & $k = 10$ & $k = 100$ & $k = 1$ & $k = 10$ & $k = 100$ & & & All & Correct & Top-1 \\
        \midrule
        \multicolumn{12}{l}{\textbf{Naive Models}} \\
        Copy
            & 0.00\% & \NA & \NAR
            & 3.32\% & \NA & \NAR
            & 90.80 & 0.00\%
            & 0.00 (0.00) & \NA & \NA
            \\
        Retrieval
            & 100.00\% & \NA & \NAR
            & 100.00\% & \NA & \NAR
            & 66.25 & 8.46\%
            & 37.50 (51.78) & 37.50 (51.78) & 37.50 (51.78)
            \\
        \specialrule{0.1pt}{2pt}{2pt}
        \multicolumn{12}{l}{\textbf{Language Models}} \\
        Seq2Seq
            & 38.36\% & 52.43\% & 62.58\%
            & 74.51\% & 89.42\% & 95.24\%
            & 88.67 & 24.35\%
            & 13.36 (25.27) & \bn{7.10 (8.16)} & 7.22 (9.32)
            \\
        CodeT5
            & \bn{72.52\%} & \bn{88.34\%} & \bn{91.95\%}
            & \bn{97.78\%} & \bn{99.49\%} & \bn{99.76\%}
            & \bn{99.64} & \bn{44.48\%}
            & \bn{8.54 (11.70)} & 8.31 (10.28) & \bn{6.84 (8.69)}
            \\
        \bottomrule
    \end{tabular*}\\[-8pt]
\end{table*}
\setlength\tabcolsep{6pt}

\subsection{Training Results}
\label{sec:results:training}

For training and fine-tuning, we adopt an early-stopping strategy to avoid overfitting, where the patience for BLEU is set to 5.
The training or fine-tuning stops when the validation BLEU is not improved for 5 epochs.

\begin{figure}
    \captionsetup{belowskip=-12pt}
    \centerline{\includegraphics[width=.85\linewidth]{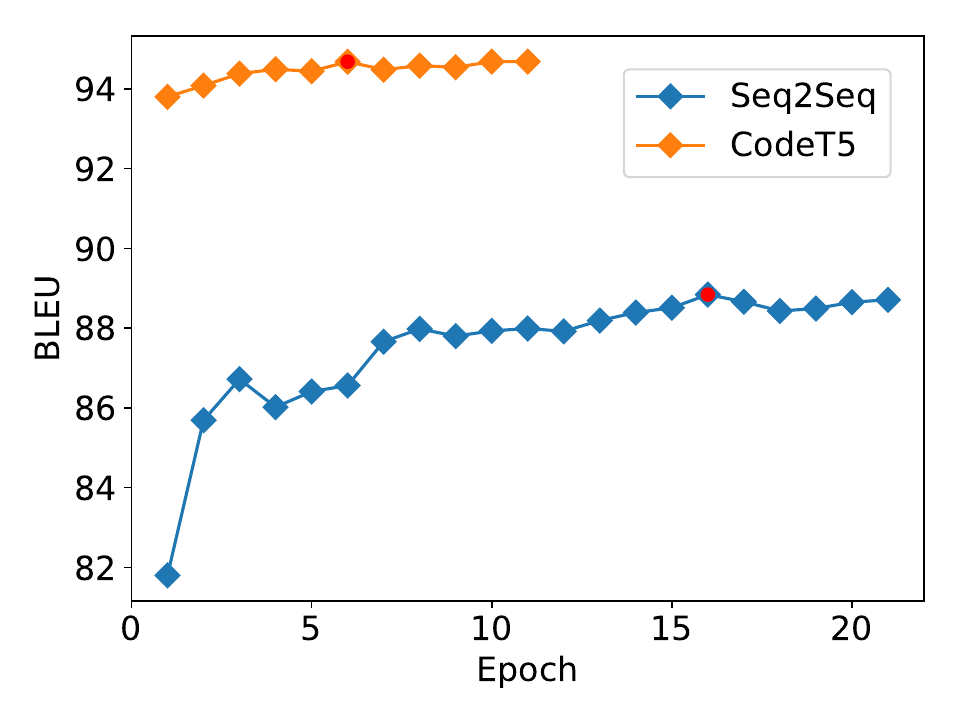}}
    \caption{\textbf{Valid BLEU throughout training/fine-tuning.} The best BLEU score for each model is represented in red point.}
    \label{fig:valid_bleu}
\end{figure}

Figure~\ref{fig:valid_bleu} shows the BLEU scores on validation set throughout training or fine-tuning.
The training takes 21.0 hours for 21 epochs in the Seq2Seq model, and the fine-tuning takes 11.5 hours for 11 epochs in the CodeT5 model.
The best validation BLEU for Seq2Seq is 88.84 at epoch 16, and that for CodeT5 is 94.68 at epoch 6.

\subsection{Evaluation Results}
\label{sec:results:evaluation}

Table~\ref{table:results} shows the evaluation results of the test set.
For sampling in Seq2Seq and CodeT5 models, we generate $n = 100$ samples for each code pair, where the sampling temperature is set to $\mathcal{T} = 0.7$, and the number of maximum tokens is set to 256.
Note that pass@$k$ of Naive Copy is ensured to be 0\% since it copies the wrong program, whereas the compilable@$k$ can be greater than 0\% since it contains \textit{the wrong but compilable program}, such as the output value is wrong.
Similarly, pass@$k$ and compilable@$k$ of Naive Retrieval are ensured to be 100\% since the Naive Retrieval retrieves the correct program from the training data.
Since the naive models copy or retrieve only one program, we only report pass@$k$ and compilable@$k$ at $k = 1$.
In addition, since the Naive Copy cannot generate any correct programs, the edit distance of correct and top-1 is invalid, and the cells are in NA.

Fine-tuned CodeT5 model performs the best on pass@$k$ and compilable@$k$ at any $k$, compared with the Seq2Seq model.
The pass@100 of 91.95\% by CodeT5 indicates that the fine-tuned CodeT5 can generate at least one correct program for 91.95\% of wrong programs by generating 100 candidates, whereas the Seq2Seq can do it for only 62.58\% of the programs.

In addition, CodeT5 performs the best on BLEU and Exact Match, compared with all baseline models.
Although the higher BLEU and Exact Match do not necessarily indicate the performance in generating better correct programs, it shows the better understanding capability of CodeT5 in generating the target programs.

For the edit distance, Seq2Seq and CodeT5 achieve 13.36 ($\pm$ 25.27) and 8.54 ($\pm$ 11.70), respectively.
As the edit distance of collected code pairs is 10.76 ($\pm$ 13.91), as shown in Table~\ref{table:code_pairs}, it shows that CodeT5 performs better than humans in repairing wrong programs with minimal edits, whereas Seq2Seq performs worse than humans.

\section{Discussion}
\label{sec:discussion}

\paragraph{Naive Copy achieves a high BLEU score}
Naive Copy achieves a BLEU score of 90.80, outperforming the Seq2Seq of 88.67, although the Exact Match is 0.00\%.
This is because, as the edit distance of the collected code pairs is small, as shown in Table~\ref{table:code_pairs}, the program repair task does not require many edits, i.e., there is already much duplication in tokens between wrong and correct programs.
Therefore, the BLEU score, which is calculated based on the matching degree of tokens between the wrong and correct programs, is likely to be high.
Although the Naive Copy cannot program repair at all, it achieves a high BLEU score due to the nature of the calculation of BLEU.
However, the fact that Seq2Seq is much worse than the Naive Copy on BLEU indicates that Seq2Seq makes unnecessary edits and makes the source program further away from the target programs.

\paragraph{Naive Retrieval achieves 100\% in pass@1}
As we mentioned, pass@1 and compilable@1 of Naive Retrieval are ensured to be 100\% since the Naive Retrieval retrieves the correct program from the training data.
Although the Naive Retrieval achieves 100\% in pass@1, the edit distance is 37.50 ($\pm$ 51.78), and it is more than three times higher than those of the collected code pairs of 10.76 ($\pm$ 13.91).
It presents a problem of the Naive Retrieval of making extensive edits, e.g., some of the input programs are converted into a completely different program if there is no similar program in the training data to be corrected.
However, although the edit distance can be large, the Naive Retrieval is still helpful in suggesting the correct program when the LMs, such as CodeT5, fail to generate the correct program.
Therefore, Naive Retrieval can be used as a hybrid with machine learning models. 

\begin{figure}
    \centering
    \begin{subfigure}[b]{.493\linewidth}
        \centering
        \includegraphics[width=\linewidth]{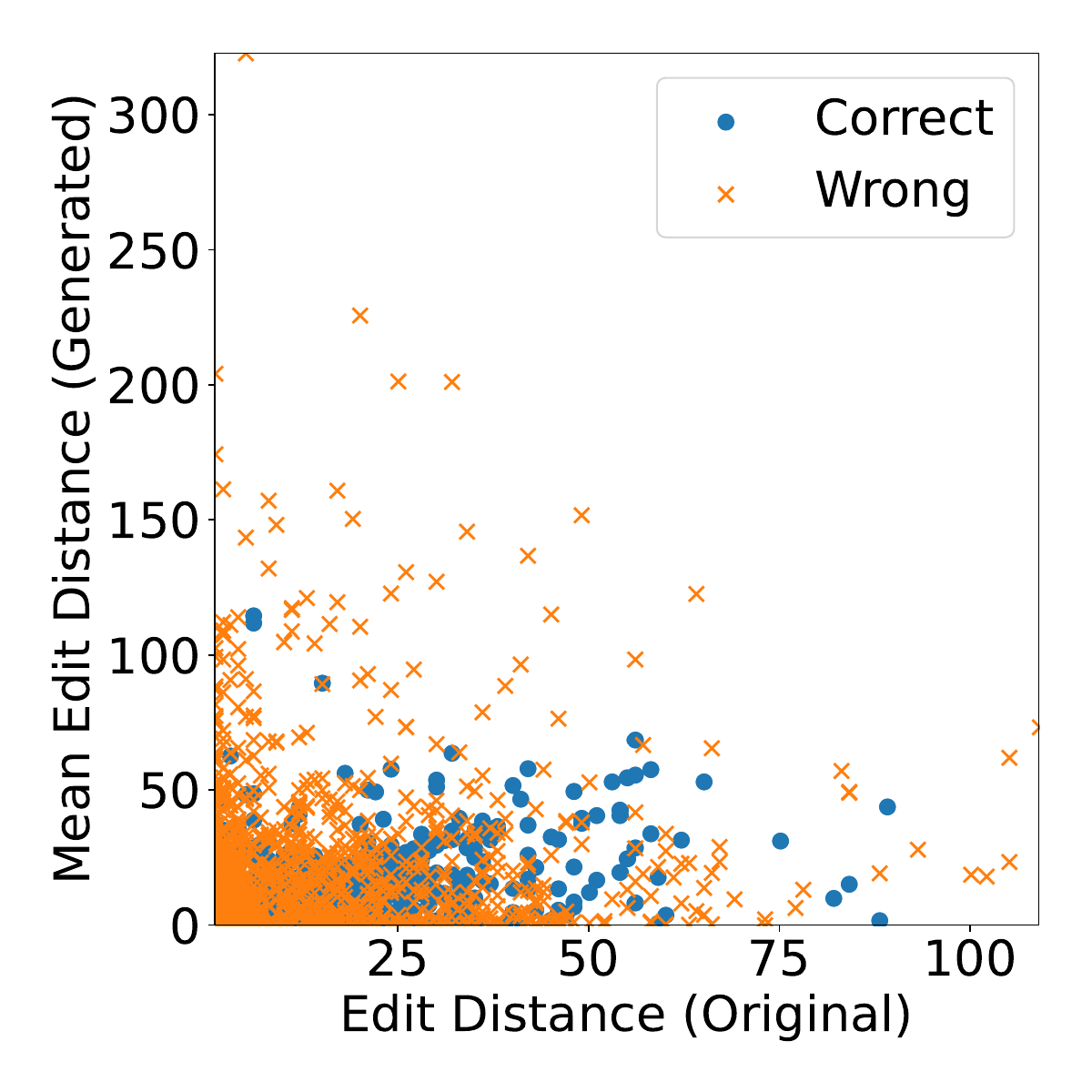}
        \caption{Seq2Seq}
        \label{fig:dist_seq2seq}
    \end{subfigure}
    \begin{subfigure}[b]{.493\linewidth}
        \centering
        \includegraphics[width=\linewidth]{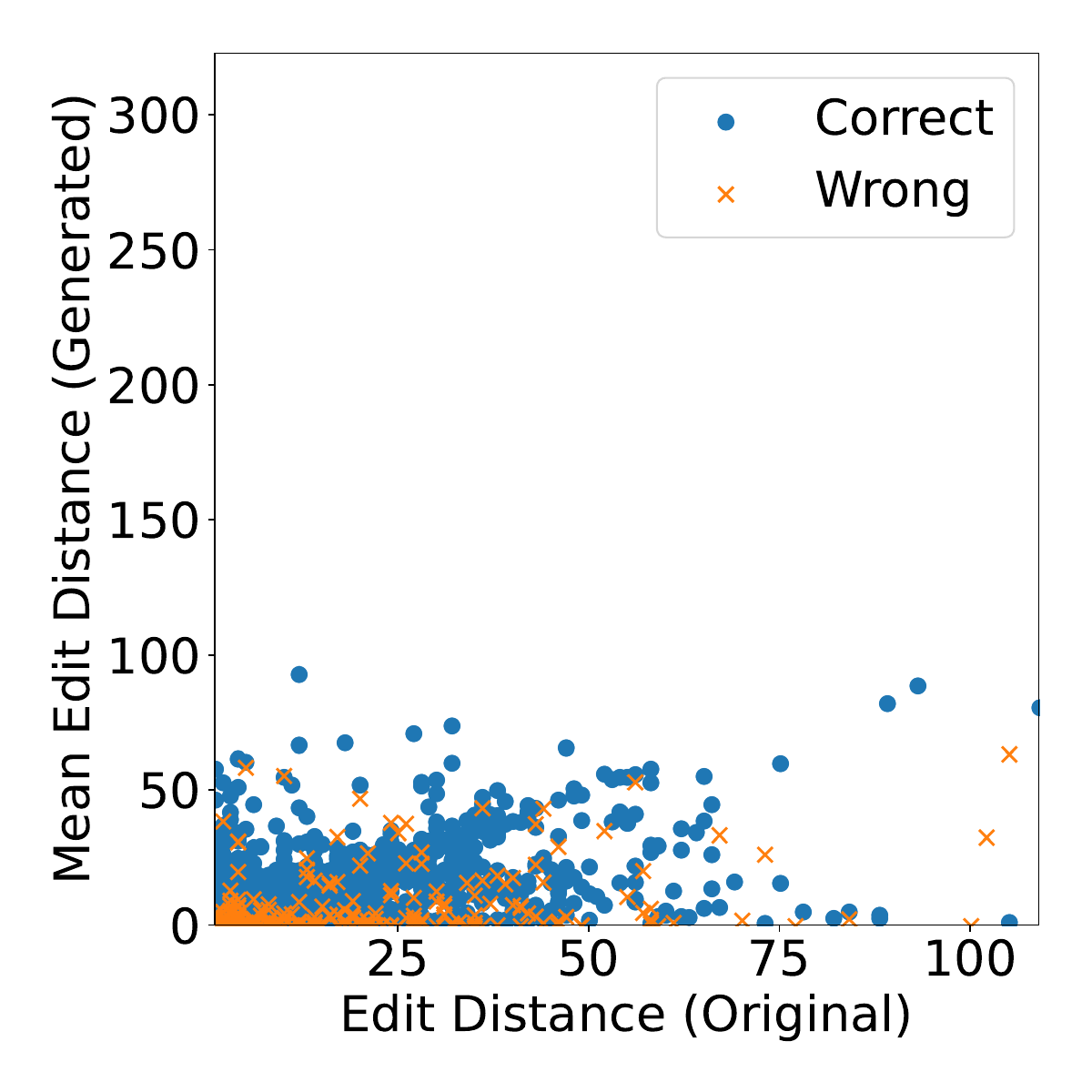}
        \caption{CodeT5}
        \label{fig:dist_codet5}
    \end{subfigure}
    \captionsetup{belowskip=-12pt}
    \caption{\textbf{Edit distance of original v.s. generated programs.} The code pair is considered \textit{correct} if at least one generated program is correct.}
    \label{fig:dist_orig_vs_gen}
\end{figure}

\begin{figure}
    \captionsetup{belowskip=-12pt}
    \centerline{\includegraphics[width=.9\linewidth]{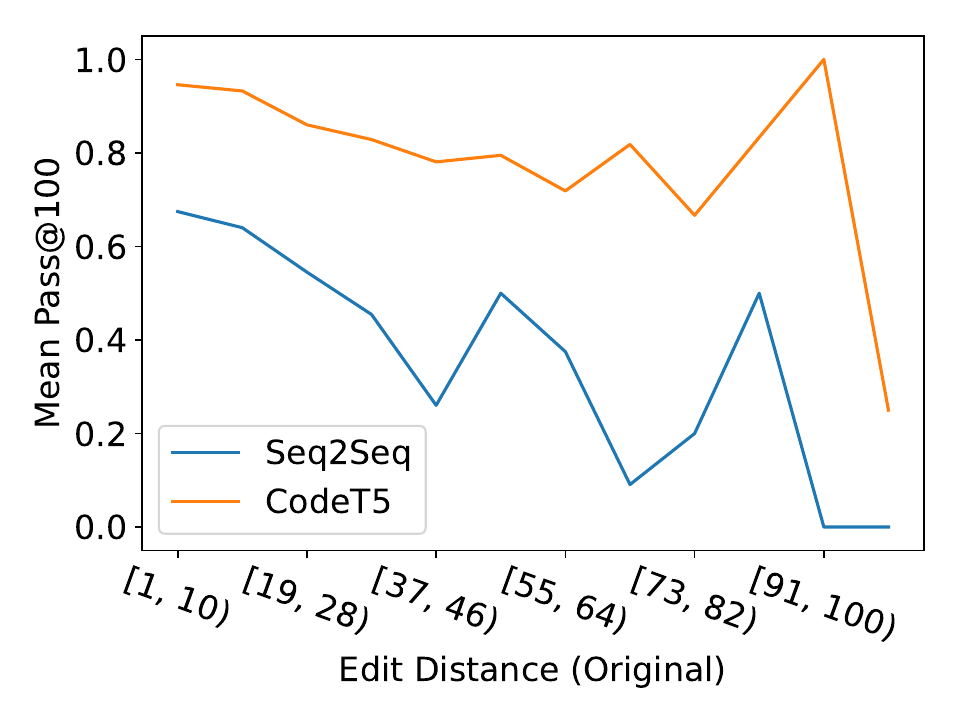}}
    \caption{\textbf{Mean pass@100 grouped by original edit distance, which is calculated between source and target programs.}}
    \label{fig:mean_pass_100}
\end{figure}

\paragraph{Edit distance of correct programs is shorter in Seq2Seq than in CodeT5}
Seq2Seq achieves 7.10 ($\pm$ 8.16) on the edit distance of correctly generated programs, and it outperforms those of CodeT5 at 8.31 ($\pm$ 10.28).
From this result, it seems that Seq2Seq performs better in generating programs with shorter edits for correct programs than CodeT5.
However, this result is strongly affected by the bias that \textbf{Seq2Seq fails to generate correct programs for the wrong programs that require longer edits}, and it does not indicate that Seq2Seq is more capable than CodeT5.
As shown in Figure~\ref{fig:dist_orig_vs_gen}, the edit distance of the generated programs by Seq2Seq (Figure~\ref{fig:dist_seq2seq}) is much larger than CodeT5 (Figure~\ref{fig:dist_codet5}).
In the figure, the points on the right indicate the programs that require longer edits, and the points on the top indicate the generated programs that have longer edits.
Therefore, Figure~\ref{fig:dist_orig_vs_gen} indicates that (1) incorrectly generated programs have longer edits in Seq2Seq, and (2) more incorrectly generated programs for programs requiring longer edits in Seq2Seq.
In addition, Figure~\ref{fig:mean_pass_100} shows that CodeT5 can generate correct programs for programs requiring longer edit distance, whereas Seq2Seq fails.

\section{Conclusion}
\label{sec:conclusion}

In this paper, we propose using a language model trained on source code, CodeT5, to suggest correct programs with minimal edits for program repair.
We fine-tune the pre-trained CodeT5 model on code pairs consisting of wrong and correct programs and evaluate its performance compared to several baseline models.
Our experimental results show that the fine-tuned CodeT5 model outperforms the baseline models in generating correct programs with shorter edit distances from the input programs.
While the naive retrieval model achieved 100\% correctness in suggesting code repairs, the average edit distance between the suggested programs and the input programs was 36.37, which is much longer than the average edit distance of 11.34 for the test data.
On the other hand, the CodeT5 model achieves a pass@100 of 91.95\%, which indicates that at least one correct program can be suggested by generating 100 candidate programs, and an average edit distance of 8.54 between the input and the generated programs, demonstrating its effectiveness in suggesting concise program repairs.
While the Seq2Seq model seems to generate programs with a shorter average edit distance than the CodeT5 model when filtering only the correct programs, it fundamentally fails to generate correct programs for incorrect programs that require longer edits.
In conclusion, the proposed method using CodeT5 shows promise in program repair by suggesting accurate and concise program repairs with minimal edits for solving introductory programming problems.
Future work includes further improvements of the correctness and edit distance and exploring other LM-based approaches for program repair.

\bibliographystyle{ieeetr}
\bibliography{main}

\end{document}